\documentclass{article}

\usepackage{amsmath,amsfonts,amssymb,times,graphicx,natbib,algorithm,algorithmic,hyperref}

\usepackage[accepted]{data4good2016}

\icmltitlerunning{Formal Concept Analysis of Rodent Carriers of Zoonotic Disease}

\begin{document}

\twocolumn[
\icmltitle{Formal Concept Analysis of Rodent Carriers of Zoonotic Disease}

\icmlauthor{Roman Ilin}{rilin325@gmail.com}
\icmladdress{Sensors Directorate,
            Bldg 620., Wright Patterson AFB, OH 45433 USA}
\icmlauthor{Barbara A. Han}{hanb@caryinstitute.org}
\icmladdress{Cary Institute of Ecosystem Studies,
           Box AB Millbrook, NY.  USA}

\vskip 0.3in
]

\begin{abstract}
The technique of Formal Concept Analysis is applied to a dataset describing the traits of rodents, with the goal of identifying zoonotic disease carriers, or those species carrying infections that can spillover to cause human disease. The concepts identified among these species together provide rules-of-thumb about the intrinsic biological features of rodents that carry zoonotic diseases, and offer utility for better targeting field surveillance efforts in the search for novel disease carriers in the wild.
\end{abstract}

\section{Introduction}
\label{sec:intro}

%Machine learning algorithms and data science techniques can help uplift humanity \cite{andrews2009close, andrews2015,DaveyPriestley2002,GanterWille1999,Han2015,Jones2009,kuznetsov2007reducing,ridgeway2006generalized,stumme2002efficient,GaliciaSW2016}.  

Formal Concept Analysis (FCA) is a data mining technique built on the foundation of lattice theory \cite{GanterWille1999}. FCA is applied to data tables with binary features (contexts) resulting in a hierarchical representation of the data (concept lattice) consisting of frequent patterns of feature co-occurrence, called concepts. Due to recent computing and algorithmic advances, it is now possible to compute concept lattices for real world data sets \cite{andrews2009close, andrews2015}. Here, FCA was applied to contexts of wild rodents to identify concepts indicative of zoonotic disease carriers, or those species carrying infections that can spillover to cause human disease. The concepts identified among these species together provide rules-of-thumb about the intrinsic biological features of rodents that carry zoonotic diseases, and offer utility for better targeting field surveillance efforts in the search for novel disease carriers in the wild.

This work builds on the analysis presented in \cite{Han2015} where a machine learning technique was applied in order to build a predictive model of carrier species. The dataset consisted of representatives of positive (carrier) and negative (non-carrier) species. The generalized boosted regression analysis built a classifier and simultaneously identified the top predictive features. The overall classification accuracy was 90%. 

Although this method was able to identify important predictive features across all species, further analysis was necessary to understand the interactions between these features, and to identify motifs of shared features that are common among subsets of positive species. An approach to such analysis is given in this contribution by utilizing FCA. FCA was conducted on binarized positive and negative contexts. We were able to identify particular biological concepts shared among rodents that carry zoonotic diseases. FCA also identified particular features for which additional empirical data collection would disproportionately improve the capacity to predict novel disease reservoirs, illustrating the kind of discourse between data mining and empirical data collection that will benefit infectious disease surveillance.

The rest of this paper is organized as follows. Section 2 provides background for FCA. Section 3 describes the rodent data and Random Forest results. Section 4 describes the application of FCA. We discuss the results and outline future plans in Section 5.

\section{Formal Concept Analysis}
\label{sec:FCA}

%Table \ref{table:numbers} and Figure \ref{fig:graph} shows.

Formal Concept Analysis (FCA) is a data mining technique built on the foundation of lattice theory. This section provides a brief mathematical description. See \cite{DaveyPriestley2002,GanterWille1999} for more details.

Suppose that $U$ is a finite set of objects, and $V$ is a finite set of properties (also called attributes or features). Suppose that a binary relation $R \subseteq U \times V $ is defined. The expression $xRy$ means that object $x$ has feature $y$ . The triplet $K=(U,V,R)$ is called a formal context.

We can express the set of features for an object $x$, as the set $xR \equiv {y \in V | xRy}$. Similarly, the set of objects with some property $y$ is the set $R y \equiv {x \in U | xRy}$. 

The following two mappings between the sets U and V are defined:

\begin{align}
A^* \equiv \bigcap_{x \in A} xR, A \subseteq U \\
B^* \equiv \bigcap_{y \in B} Ry, B \subseteq U 
\end{align}

$A^*$ is the set of features shared by all objects in the set $A$. Similarly, $B^*$ is the set of objects that all share the same set of features. An ordered pair $(A,B), A \subseteq U, B \subseteq V$, is called a \textbf{ formal concept }, if $A^*=B$ and $B^*=A$. The set of objects in the formal concept is referred to as the \textbf{extension}, and the set of features as the \textbf{intension} of the concept. For a concept $C$, the extension and the intension are denoted as $Ext(C)$ and $Int(C)$. It is a classical result in FCA that the mappings between $U$ and $V$ form Galois connection.  

The mappings

\begin{align}
X^{**}: 2^U \rightarrow 2^U \\
Y^{**}: 2^V \rightarrow 2^V 
\end{align}

are closure operators. A fixpoint of the closure operator, which is a set such that $X^{**}=X$, is called a closed element. As evident from the definition of formal concept, the set of intensions of all concepts is equivalent to the set of all elements closed with respect to the mapping (3), and similarly, the set of all extensions is the set of closed sets with respect to (4). We denote the set of all concepts of a given context as

\begin{equation}
\boldsymbol{B} \equiv  \left\{ (A,B)│A^*=B  \wedge B^*=A \right\}
\end{equation}

and the sets of all intensions and extensions as  $INT(\boldsymbol{B})$ and $EXT(\boldsymbol{B})$.

Consider the set B of all concepts of a given context. We define a relation of partial order of this set, as follows: 

\begin{equation}
(A_1,B_1) \preceq (A_2, B_2) \equiv A_1 \subseteq A_2 \equiv B_1 \supseteq B_2
\end{equation}

This relation defines a partially ordered set that is a complete lattice \cite{DaveyPriestley2002}, referred to as \textbf{concept lattice}. Suppose that $T$ is a set of indices. The meet and join of concepts from the set $(A_t, B_t ), t \in T$, are defined as follows:

\begin{align}
\bigwedge_{t \in T} (A_t, B_t) \equiv \Big(  \big( \bigcap_{t \in T}  A_t \big) ,  \big( \bigcup_{t \in T} B_t  \big)^{**}  \Big) \\
\bigvee_{t  \in T} (A_t, B_t) \equiv \Big(  \big( \bigcup_{t \in T} A_t  \big)^{**} ,   \big( \bigcap_{t \in T}  B_t \big)   \Big)
\end{align}

Note that the intersection of concept extensions is an extension of some concept, but the union of extensions is not necessarily an extension of some concept. The same is true for intensions: the intersection of intensions is an intension of some concept, but not the union. We will denote the transformation of a context into the concept lattice as

\begin{equation}
\boldsymbol{B} \equiv CL(K)
\end{equation}

An illustration of FCA is shown in Fig. 1 for a simple context. The diagram was computed using a software package called Galicia \cite{GaliciaSW2016}. The lattice diagram is drawn by placing the more generic concepts (those containing fewer features) above more specific concepts (with more features).

\begin{figure*}[t]
\vskip 0.2in
\begin{center}
\centerline{\includegraphics[width=\textwidth]{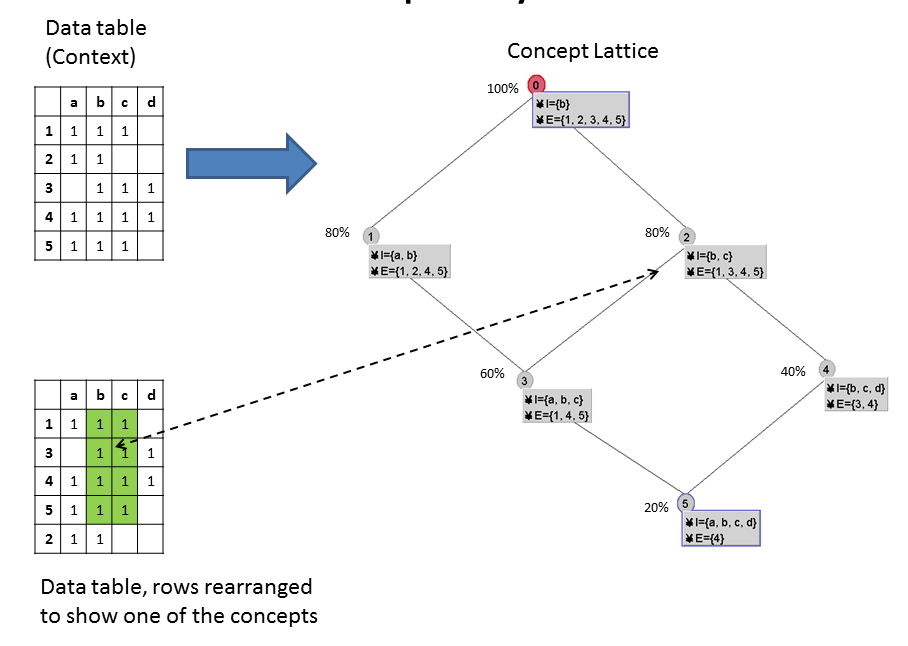}}
\caption{Illustration of Formal Concept Analysis. Data table contains 5 objects and 4 attributes. The concept lattice consists of 6 concepts. A concept can be thought of as a square in the data table, which is illustrated for one of the concepts.}
\label{fig:fca}
\end{center}
\vskip -0.2in
\end{figure*}

Each concept is characterized by a number between 0% and 100% called support. This is defined as the number of objects in the context supporting given concept, or formally, as

\begin{equation}
Support(C) \equiv \frac{|Ext(C)|}{|U|} \times 100 \%
\end{equation}

Fig. 1 shows the support for each concept in a simple context. The greater the support, the more prominent a given concept. Also, it is easy to show that support increases as we move up in the lattice, with support of the top concept being 100\%. This property led to the development of an Iceberg lattice (Stumme, 2002). For a given concept lattice, the Iceberg lattice contains only the concepts with support above a certain threshold. This is equivalent to cutting off the bottom of the lattice. The remaining elements contain more generic concepts that explain a large portion of the context. In the example in Fig. 1 the 80\% Iceberg would contain the top two concepts (disregarding the trivial top concept). We observe that 80\% of all the objects contain one or both of these concepts.

\section{Rodent Data}
\label{sec:data}

Rodent trait data were obtained from PanTHERIA, a species-level database of life history, ecological, and geographical traits of the world’s mammals \cite{Jones2009}. The traits include features such as adult body mass, age of first birth, litter size; geographical region given by the maximum and the minimum latitude and longitude, etc. The total number of predictive features was 88. Values are the result of numerous field studies, thus there are many missing values in the dataset.

The rodent dataset contained 2277 objects (species), with 217 labelled as positive (disease reservoirs), and the rest as negative (reservoir status unknown). The analysis in \cite{Han2015} utilized boosted regression trees to train classifiers, using up to 10,000 trees and 10-fold cross-validation to prevent overfitting. Analysis was done using gbm package in R \cite{ridgeway2006generalized}. This analysis identified top 15 predictor features (Table 1).

\begin{table}[ht]
\caption{Top 15 predictor features for Rodent Data}
\label{table:data}
\vskip 0.15in
\begin{center}
\begin{tiny}
\begin{sc}
\setlength{\tabcolsep}{2pt}
\begin{tabular}{|c|c|}
\hline
\verb"X26.1_GR_Area_km2"  & Geographic area, square km.  \\\hline
\verb"X23.1_SexualMaturityAge_d"	& Age of sexual maturity, days \\\hline
\verb"X27.2_HuPopDen_Mean_n.km2"	& Human population density, $km^{-2}$ \\\hline
\verb"logNeoBM"  & Neonatal body mass, log(grams)  \\\hline
\verb"X15.1_LitterSize"  &	Litter size \\\hline
\verb"X5.1_AdultBodyMass_g"  &	Adult body mass, g  \\\hline
\verb"X9.1_GestationLen_d"	  & Gestation length, days  \\\hline
\verb"X25.1_WeaningAge_d"  &	Weaning age, days  \\\hline
\verb"X13.1_AdultHeadBodyLen_mm" &	Adult length head and body, mm \\\hline
\verb"SpeciesDensity"	  & Density of mammal species, $km^{-2}$  \\\hline
\verb"X30.2_PET_Mean_mm"	&  Mean potential evapotranspiration rate, mm   \\\hline
\verb"X26.2_GR_MaxLat_dd"   &	Maximum latitude of the geographical range   \\\hline
\verb"X16.1_LittersPerYear"	  &  Number of litters per year   \\\hline
\verb"X26.5_GR_MaxLong_dd"  &  	Maximum longitude of the geographical range  \\\hline
\verb"X26.3_GR_MinLat_dd"	   &  Minimum latitude of the geographical range   \\\hline
\end{tabular}
\end{sc}
\end{tiny}
\end{center}
\vskip -0.1in
\end{table}

\section{Analysis of Concepts in Rodent Data}
\label{sec:analysis}

\begin{figure*}[t]
\vskip 0.2in
\begin{center}
\centerline{\includegraphics[width=\textwidth]{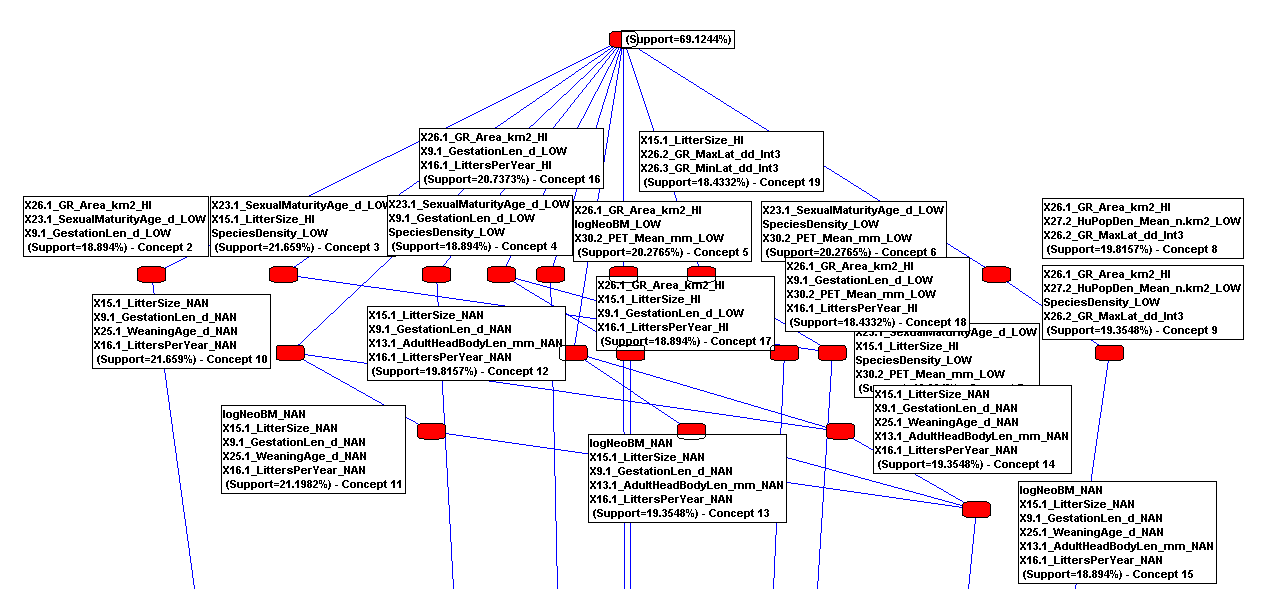}}
\caption{Iceberg Lattice for Rodent Context at 18\% minimum support.}
\label{fig:fca}
\end{center}
\vskip -0.2in
\end{figure*} 

Formal Concept Analysis was conducted on the reduced dataset containing 2277 objects and 15 features. Since the features are integer or real valued, they had to be discretized. For non-geographical features, we computed the median values, and mapped each value to one of three categories: high (above median), low (below median), and NAN (missing). For geographical features, latitude was separated into 4 bins: NAN (missing), [-90, -30], [-30, 30], and [30, 90].  Longitude was separated into 3 bins: NAN (missing), [-180, -25], and [25, 180]. Biogeographical boundaries roughly correspond to the tropical vs. subtropical regions and Eurasia vs. the Americas.  The resulting binary dataset contained 47 columns, one for each feature category.

Formal Concept Analysis was conducted using In-Close software \cite{andrews2009close}. The positive context, with 217 objects, resulted in 6,197 concepts. The negative concepts, with 2,060 objects, resulted in 137,515 objects. Large number of concepts is typical for FCA. Iceberg analysis was applied to the large number of concepts, which is typical of FCA, to focus on more general concepts present in the data and identify strong dependencies between features.

Each concept is a pattern of feature co-occurrence in the data. If a certain pattern is found in both the negative and the positive lattice, it alone cannot be used to decide whether a certain object belongs to a positive or negative class. Based on this reasoning, we removed all concepts from the positive lattice that also exist in the negative class. The reduced positive lattice with 738 concepts was used for further analysis.
We computed the Iceberg lattice with minimum support of 18\%. The diagram is shown in Fig. 2. This lattice has 18 concepts, and 6 of them are the “missing data” concepts – those with names ending in NAN.  These concepts are supported by 69\% of the objects in the positive class.

\section{Conclusion}
\label{sec:conclusion}

This contribution outlines the initial results of applying FCA to the rodent data. We provided a brief description of FCA as a mathematical framework for frequent pattern search in categorical datasets. 
FCA improves existing methods for identifying traits characterizing potential zoonotic disease carriers by exploring the dependencies between different traits. FCA also identifies particular data items that are disproportionately represented among disease carriers that should be prioritized for future field work. For example, we found that about 21\% of all positive species contain a pattern of large litter size, early age at sexual maturity, and living in areas with high mammal biodiversity (species density). This pattern is not encountered in the negative species, suggesting that more species with this pattern need to be tested for zoonotic diseases. We also found that about 21\% of positive species contained a pattern of missing data on weaning age, litter size, length of gestation period, and the number of litters per year. However, these data were available for negative species. This contrast highlights a particular need to measure this suite of features over others in future fieldwork to make the greatest improvements in the prediction of positive species.
In future work, we plan to expand the analysis by utilizing more concepts from FCA. For example, we plan to evaluate stability scores of concepts \cite{kuznetsov2007reducing}. We also plan to expand our analysis to other datasets, in particular to the analysis of bats to identify concepts describing species that carry filoviruses, which cause hemorrhagic fevers such as Ebola virus disease.

\bibliographystyle{icml2016}
\bibliography{testlib}

\end{document}